\pdfoutput=1

\documentclass[runningheads]{llncs}

\usepackage{amssymb}
\usepackage{graphicx}
\usepackage{subfigure}
\usepackage{epstopdf}

\usepackage{url}
\usepackage{amsmath}
\usepackage{array,multirow}

\urldef{\mailsa}\path|{marjan.alirezaie, amy.loutfi}@oru.se|  
\newcommand{\keywords}[1]{\par\addvspace\baselineskip
\noindent\keywordname\enspace\ignorespaces#1}

\begin{document}
\mainmatter  

\title{Reasoning for Improved Sensor Data Interpretation in a Smart Home}


%
%
\author{Marjan Alirezaie%
\and Amy Loutfi}
%

\institute{Center for Applied Autonomous Sensor Systems (AASS), \\
Dept. of Science and Technology,\\
\"{O}rebro University, SE-701 82, \"{O}rebro, Sweden\\
\mailsa\\
}

%
%

\maketitle
\setcounter{page}{1}
\thispagestyle{plain}

\begin{abstract}
In this paper an ontological representation and reasoning paradigm has been proposed for interpretation of time-series signals. The signals come from sensors observing a smart environment. The signal chosen for the annotation process is a set of unintuitive and complex gas sensor data. The ontology of this paradigm is inspired form the SSN ontology (Semantic Sensor Network) and used for representation of both the sensor data and the contextual information. The interpretation process is mainly done by an incremental ASP solver which as input receives a logic program that is generated from the contents of the ontology. The contextual information together with high level domain knowledge given in the ontology are used to infer explanations (answer sets) for changes in the ambient air detected by the gas sensors.

\end{abstract}

\section{Introduction}
\label{sec:intro}

Due to the advances in sensor technology which have resulted in affordable and easy-to-use sensors, a large segment of research has
often been focused on integration of sensors in other intelligent systems. Such systems may consist of a mobile system, such as a robot or a  distributed system of heterogeneous networked sensors, such as a smart home. Regardless of their purposes, applications of smart environments a.k.a context-awareness applications are common in continuous observation of the environment, and consequently, in issues related to the sensor observation process \cite{P40}. Therefore, the representation and reasoning model suggested for these applications need to be able to deal with the issues such as data incompleteness, lack of observation and the dynamic situation of smart environments. 

In this paper, we present a knowledge driven approach to reasoning about changes detected in the quality of the pervasive air in a smart environment. The smart environment is a kitchen equipped with a sensor network including a gas sensor which is continuously sniffing the ambient air. By modelling high level knowledge about odours, their causes, and relations to other phenomenon, it is possible to assist the interpretation of the gas sensor signals.  However, to automate this process, the high level (symbolic) knowledge needs to be seamlessly connected to the lower level sensor (quantitative) data. Furthermore suitable reasoning techniques are also required in order to infer information beyond that which is measured by the gas sensors or the sensor network. 

Automated reasoning is achieved via answer set programming (ASP). The motivation to using ASP is twofold. First, the proposed application should deal with incomplete data caused by  uncertain behaviour of sensors or the lack of observations. Second, the dynamic nature of the sensor network should be taken into account where new observations can influence the reasoning process and if necessary invalidate previously inferred results. In our proposed paradigm, a non-monotonic reasoner is advocated and ASP has the further advantage of using an incremental solver which is suitable for stream reasoning \cite{P17}. 

The necessary high level knowledge for the reasoner is modelled within an ontology inspired by the Semantic Sensor Network (SSN) \cite{M2}. The proposed ontology provides the basis for declarative definition of entities, features and events in the environment. Also, this ontology enables a reuse of information, compatible with the emerging trend of  linking contextual information to sensor data seen in the Internet of Things (IoT) paradigm. 

The goal of the proposed knowledge representation and reasoning paradigm is important given the trend IoT towards a global connectivity between objects from the physical world. As more sensors of varying modality become connected, it will be of importance to provide automated interpretation of the sensor data and to exploit existing resources to this aim if possible. However, the integration of ontology languages such as OWL-DL with non-monotonic reasoning is non-trivial. Therefore a contribution in this work is to implement a conversion between OWL-DL ontologies and the incremental ASP reasoner.

This paper is organized as follows. In Section~\ref{sec:relatedWorx} we study the related work. The essential components required for implementation of a smart environment are explained in Section~\ref{sec:architecture}. Section~\ref{sec:reasoningOdours} describes the details of the reasoning process. The evaluation results of the proposed approach are discussed in Section~\ref{sec:results} which is continued by the conclusion.

\section{Related Work}
\label{sec:relatedWorx}

The relevant subset of works such as \cite{P10} deal with semantic web technologies, and the areas of the semantic sensor web. These works propose to semantically annotate sensor data with semantic annotations published on the semantic Web (Semantic Sensor Web). In this way, interoperability between sensor networks and consequently the situation awareness increases 
 In this domain, ontologies are the most popular method to achieve such structures. For example, the Semantic Sensor Network (SSN) ontology is a suggestion from the W3C Semantic Sensor Networks Incubator Group (SSN-XG) \cite{M2} for representing the sensor data independent of the domain knowledge. The concentration of the SSN ontology is on sensors (and their capabilities), observations, and systems rather than the domain-dependent concepts. 
 
Due to the expressivity of the ontology languages such as OWL-DL, definition of concepts, relations and axioms in the domain knowledge are declarative. In order to reason upon this knowledge, therefore, an expressive reasoning technique is required. All the deductive (monotonic) reasoners based on description logic (DL) implemented for ontologies, depending on their formalisations, provide different degrees of complexity and expressiveness. Nevertheless, applications need to deal with the issue of incomplete data caused by inherently uncertain behaviour of sensors or the lack of observation. Furthermore, the dynamic behaviour of sensory systems where the newly added observation can influence the reasoning process such that the new reasoning outputs may contradict the previously inferred results, indicates the need for non-monotonic reasoning techniques. Recently, hybrid approaches to combining monotonic and non-monotonic reasoning have been proposed.  For example, \cite{A15} combine a handmade ontology representing high level concepts and the ASP solver as a non-monotonic reasoner to reason over sensor data. These works illustrate the potential of ASP to cope with stream reasoning. In this paper, in addition to using an incremental ASP solver \cite{P17}, we utilize an ontology synonymous to the SSN ontology and provide the means to automatically generate the logic used by the solver. 

\section{System Architecture}
\label{sec:architecture}

This section provides a high level view of the system architecture depicted in Fig.~\ref{fig:framework}. The input is the sensor data from the sensor network and the output is the annotations that consist of the reasoner's explanations to changes detected in the gas sensor (target sensor). The system architecture consists of three blocks: \textit{A) Observation Process}, \textit{B) Data/Knowledge Integration} and \textit{C)Reasoning} that are described in the following sections.

\subsection{Block A : Observation Process}
\label{sec:sensorNetwork}

\begin{figure*}[t]
  \centering
    \includegraphics[width=11cm,height=5cm]{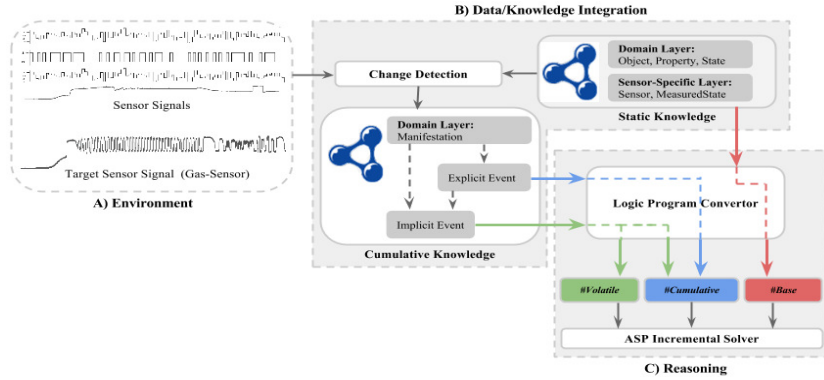}
    \caption{Knowledge Representation and Reasoning Framework.}
    \label{fig:framework}
\end{figure*}

The observation process of a smart home is performed through a set of heterogeneous sensors which are synchronously and continuously observing the environment. Each object in the smart home, depending on its attributes of interest can be observed by one or several sensors. Sensors include motion detectors, luminosity sensors, temperature sensors, magnetic contacts etc. In order to lighten the data load, instead of continuously sampling of data, we chose the event capturing approach which is accomplished by passing the data stream through the change detection process (see Fig.~\ref{fig:framework}). In Section~\ref{sec:cumulative}, we show how each data point related to a change in the signal is represented in form of a \textit{manifestation} considered as an event. Although the data output of an event capturing approach is not as complete as the data in the continuous data sampling approach, it is possible to cope with gaps in the sensor output via ASP in handling incompleteness of data. 

\subsection{Block B: Data/Knowledge Integration}

Both time dependent and independent concepts are modelled within the processes in Block \textit{B}. The static knowledge which represents the time independent concepts are related to the observed phenomena and their properties. The cumulative knowledge represents the time dependent concepts related to observations in the sensor data and events. For the sake of facilitating the transformation from the ontology  to the solvable logic program by the incremental ASP solver, we separate the representation of static knowledge from the cumulative knowledge.

\subsubsection{Static Knowledge}
\label{sec:staticknow}
\begin{figure*}[t]
  \centering
    \includegraphics[width=\textwidth]{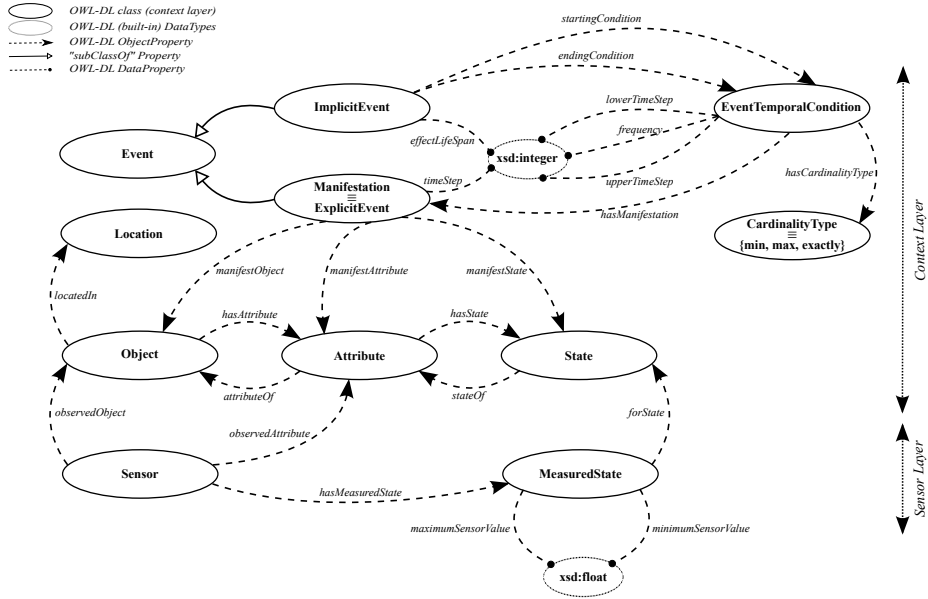}
    \caption{The hierarchical structure of the OWL-DL Ontology}
    \label{fig:upper-ontology}
\end{figure*}

The ontology depicted in Fig.~\ref{fig:upper-ontology} is made up of two layers, namely a context layer and a sensor layer. The context layer specifies general facts about the context along with their constraints, whereas the sensor layer contains concepts with quantitative values related to sensors.

\emph{\textbf{Context Layer:}} The class \textit{Object}, in the ontology refers to all entities that can be a source of an event (detected by the target sensor) in the environment. For instance: a freezer is considered as an object (\textit{Freezer} $\sqsubseteq $\textit{Object}), since its stuff inside can be rotten and smell. Each object is defined based on a set of attributes such as temperature (\textit{Temperature} $\sqsubseteq $ \textit{Attribute}), illumination  or electric-current, whose situations are significant in event definition. The situation of attributes, furthermore, are defined as individuals of the class \textit{State} which independent of the sensor values, represent all possible situations (e.g., \textit{cold}, \textit{warm}, \textit{dark}) of an attribute. As mentioned before, there is at least one sensor in the environment (e.g., a gas sensor) whose data is targeted for the interpretation process. The difference between modelling the target sensor data and the others is in definition of their states. For the target sensor, the possible states are limited into two \textit{normal} and \textit{abnormal} states. Once the sensor data is out of the value range of the \textit{normal} state, the state is switched into \textit{abnormal} which needs to be interpreted. 

\emph{\textbf{Sensor Layer:}} The sensor layer (Fig.~\ref{fig:upper-ontology}), contains concepts representing quantitative sensor-related data. The class \textit{Sensor} is responsible for holding sensors' information used for the observation process. Individuals of the class \textit{State} (e.g., \textit{cold}, \textit{warm}) symbolically explain objects in terms of their attributes' situations. However, since it is the sensor output which shows the state of the observed object, we need to create a link between the symbolic states and the real observation values. For this, we define the \textit{MeasuredState} class whose individuals relate sensor values to specific states. Each individual of the \textit{MeasuredState} class has two properties, \textit{minimumSensorValue} and \textit{maximumSensorValue}, providing a range of values for a state. For example, the triple \textit{(min=1, max=3, forState=cold)} creates a \textit{MeasuredState} instance which is related to a particular sensor via the \textit{hasMeasuredState} property.

\subsubsection{Cumulative Knowledge}
\label{sec:cumulative}
The cumulative knowledge takes events captured by the change detection and translates them into symbolic concepts that either denote \textit{ExplicitEvents}  called \textit{Manifestations} or \textit{ImplicitEvents}. An explicit event is one which is directly measured by the sensors. For instance, whenever the temperature of the freezer changes from a cold state to a warm state, the change detection component generates a manifestation such as \textit{m:(freezer, temperature, cold, t)}, regardless of the range of sensor data for a cold freezer. The last property of the manifestation class is the \textit{timeStep} (\textit{t}) which is related to the incremental step considered by the reasoner. The subclasses of the \textit{Manifestation} class are generated by literally combining the the name of object, attribute and state concepts involved in the event. For example, a tentative manifestation can be \textit{FreezerTemperatureWarm} which indicates an event related to a freezer (object) whose temperature (attribute) gets warm (state). 

The \textit{ImplicitEvent} class addresses those events that are not directly observable by sensors but  triggered based on a rule set that consists of Explicit Events. For example, in case of a kitchen, events such as  \textit{Cooking} and  \textit{Rotting} are defined as subclasses of the \textit{ImplicitEvent} class, and are the main events that cause a change in the quality of the kitchen's ambient air. 

Each \textit{ImplicitEvent} so as to be inferred by the reasoner, needs a set of specific a priori conditions to be detected in the environment. These pre-conditions are characterized as instances of the \textit{EventTemporalCondition} class. Two properties, \textit{startingCondition} and \textit{endingCondition} (Fig.~\ref{fig:upper-ontology}), relate an implicit event to its conditions which indicate the situations required for inferring the event started or ended, respectively. 
The class \textit{EventTemporalCondition} as such is related to manifestations. In other words, an event's conditions are defined based on manifestations. Therefore, the class \textit{EventTemporalCondition}, has one relation with the \textit{Manifestation} class via the \textit{hasManifestation} property. 

Moreover, the detection of an event's preconditions is expected to be during a specific time around the time step of the event. Therefore, each precondition is also assigned with two integer values indicating the lower and upper bound of a range for the event time step. The preconditions of each implicit event inferred at time step \textit{t}, can happen either at the same time step or before it. In case of the coincidence, the two integer values are set to zero, otherwise they are set to positive numbers. For example, given \textit{Rotting} as an implicit event, we define a subclass for \textit{EventTemporalCondition}, e.g., ETC1. One of the manifestations that can be related to the rotting process is \textit{FreezerTemperatureWarm}. In order to infer \textit{Rotting} at time step \textit{t}, its manifestation, for example, has to be detected between time step \textit{t-B} and \textit{t-A} where both \textit{A} and \textit{B} ($0 \leq $ \textit{A} $\leq$ \textit{B}) are integer values assigned to the class ECT1 via the \textit{upperTimeStep} and the \textit{lowerTimeStep} properties by the user, respectively. 


As shown in Fig.~\ref{fig:upper-ontology}, each implicit event is assigned to one or several \textit{EventTemporalCondition} concepts via its \textit{startingCondition} and \textit{endingCondition} properties. These relations between implicit events and their conditions are set by the user based on the features of the scenario. Further details about implicit events are given in Section~\ref{sec:reasoningOdours}.

\section{Reasoning about Odours}
\label{sec:reasoningOdours}

In this section, the details about the conversion process and the reasoning process, depicted in Block \textit{C} (Fig.~\ref{fig:framework}), are discussed based on the notations explained in previous sections. 

\subsection{Generating the Base Logic Program ($\mathcal{B}$)}
\label{sec:programB}

In order to generate the base logic program $\mathcal{B}$ which is time independent, for each individual object, attribute and state defined as the static knowledge in the ontology, two logical literals including a fact and a unary predicate are added to the program $\mathcal{B}$ as follows:

%
%
%

\begingroup
\fontsize{6pt}{7pt}\selectfont
\begin{flalign*}
& \forall \textit{o} \in \textit{O} \sqsubseteq \textit{Object} \mapsto \textit{B} = \textit{B} \cup \{\textit{o}, \textit{O(o)}\}\\
& \forall \textit{a} \in \textit{A} \sqsubseteq \textit{Attribute} \mapsto \textit{B} = \textit{B} \cup \{\textit{a}, \textit{A(a)}\}\\
& \forall \textit{s} \in \textit{State} \mapsto \textit{B} = \textit{B} \cup \{\textit{s}, \textit{State(s)}\}
\end{flalign*}
\endgroup


Therefore, the final $\mathcal{B}$ logic program will contain a set of both unary and binary grounded predicates.

\subsection{Generating the Cumulative Logic Program ($\mathcal{P}[t]$)}
\label{sec:programP}
The cumulative logic program represented with $\mathcal{P}$ is generated by time dependent concepts in the ontology. Time dependent concepts containing the time step parameter $t$ in their definitions include both the explicit (manifestation) and the implicit events.

The logic program $\mathcal{P}$ is incrementally extended meaning that whenever an explicit event is captured, a manifestation indicating an object, its attribute and its state at time step \textit{t}, is converted into an appropriate predicate and extends $\mathcal{P}$ with a new manifestation predicate as follows:

\begingroup
\fontsize{6pt}{7pt}\selectfont
  \begin{align*}
      \mathcal{P} = \mathcal{P} \cup \{ manifestation(o, a, s, t) \}.
  \end{align*}
\endgroup

Moreover, whenever a manifestation is generated, a rule based on the pattern $r_{1}$ shown in the following is added to the program $\mathcal{P}$. This rule allows the solver to infer appropriate explicit events related to the manifestation. The head of the logic rule, \textit{O\_A\_s(t)}, indicates a predicate whose name is generated by concatenating the name of the subclass of the class \textit{Object} ($O$), the name of its attribute ($A$) and the state name ($s$). The integer value $t$ of $\mathcal{P}[t]$ refers to the last parameters in a manifestation which correspond to the time step at which the change is captured. 

\begingroup
\fontsize{6pt}{7pt}\selectfont
  \begin{flalign*}
      &r_{1}: O\_A\_s(t) :- \hspace{7pt} manifestation(X, Y, s, t), \hspace{3pt} O(X), \hspace{3pt} A(Y).
  \end{flalign*}
\endgroup

Apart from explicit events, implicit events (e.g., \textit{Rotting}) also have a considerable impact on extension of the program $\mathcal{P}$. Since implicit events are not directly observable via sensors, and are defined based on temporal relations between explicit events, the time difference plays an essential role in their definition. For instance, the event \textit{Rotting} starts after passing a specific amount of time (e.g., 1 day) when the explicit event \textit{FreezerTemperatureWarm} has been detected. Therefore, for the sake of inferring implicit events and measuring the time difference between events, we need to continuously sample sensor data. 

Nevertheless, as mentioned in Section~\ref{sec:sensorNetwork}, to hinder the reasoner to be overwhelmed with data, instead of continuous data sampling, we chose the event capturing approach. Compensating the lack of continuous data sampling is achieved by defining each implicit event within three logical rules. The two first rules indicate the conditions required for inferring the starting and the ending of an implicit event, respectively. These conditions are defined in the ontology as the relations between an implicit event and manifestations. As we can see in Fig.~\ref{fig:upper-ontology}, the \textit{ImplicitEvent} class is related to the \textit{Manifestation} class via the \textit{EventTemporalCondition}. The body part of the first rule which indicates the conditions required to infer an implicit event's inception, is generated based on the conjunction of all manifestations that are connected to the instances of the \textit{EventTemporalCondition} class via the \textit{startingCondition} property. Likewise, the body part of the second rule regarding the ending conditions of an implicit event, is the conjunction of all manifestations, that are related to the \textit{EventTemporalCondition} class, however, via the \textit{endingCondition} property. For instance, assuming the following axioms are defined in the ontology, we generate the two first rules for the implicit event \textit{Garbage}:
 
For instance, the rules $r_{2}$ and $r_{3}$ related to the inception and ending of the \textit{Garbage} event will be as follows:

\begingroup
\fontsize{6pt}{8pt}\selectfont
  \begin{flalign*}
      &r_{2}: garbage(t) :- \hspace{7pt}trashBinDoorOpen(t), \hspace{3pt}trashBinIlluminationDark(t).\\
      &r_{3}: garbageEnd(t) :- \hspace{5pt}garbage(t-1), \hspace{2pt}trashBinDoorOpen(t), \hspace{2pt}trashBinIlluminationBright(t).
  \end{flalign*}
\endgroup
 
The third rule ($r_{4}$), also indicates the event progression conditions. The term ``ImplicitEvent'' in $r_{4}$ refers to a predicate equivalent to an \textit{ImplicitEvent} concept (e.g., \textit{Garbage})):

\begingroup
\fontsize{6pt}{8pt}\selectfont
  \begin{flalign*}
      &r_{4}: ImplicitEvent(t) :- \hspace{7pt} ImplicitEvent(t-1), \hspace{3pt} not \hspace{3pt} ImplicitEventEnd(t).
  \end{flalign*}
\endgroup

Implicit events which can declaratively express the ambient smell in the kitchen are used as meaningful explanations for changes detected over the gas sensor. In order to infer that the smell of an implicit event is sensed, two conditions have to hold: First, the truth of a manifestation denoting an abnormal state for the target object (\textit{airSmellAbnormal}) at the current time $t$, and second, the truth of an implicit event:

\begingroup
\fontsize{6pt}{8pt}\selectfont
  \begin{flalign*}
      &r_{5}:smellImplicitEvent(t) :- \hspace{7pt} airSmellAbnormal(t), \hspace{3pt}ImplicitEvent(t).
  \end{flalign*}
 \endgroup 
 
However, there are situations in which we cannot infer the truth of an implicit event,  its smells still stays in the environment. Due to gradually fading of smells, apart from $r_{5}$ which indicates the preliminary conditions for the \textit{smellImplicitEvent} predicate, the rule $r_{6}$ is also added to the program $\mathcal{P}$. The term ``VALUE'' refers to an integer value showing an approximate time interval during which the smell, even after ending the event, normally stays in the ambient air. This value is set in the ontology for an implicit event instance, via the \textit{effectLifeSpan} property (Fig.~\ref{fig:upper-ontology}):

\begingroup
\fontsize{6pt}{8pt}\selectfont
  \begin{flalign*} 
&r_{6}:smellImplicitEvent(t) :- \hspace{7pt} airSmellAbnormal(t), \hspace{3pt}smellImplicitEvent(t-1), \\
&\hspace{120pt}1\{ImplicitEventEnd(t-VALUE..t)\}.
  \end{flalign*}
\endgroup

In order to interpret the smell of an inferred implicit event and explain the current state of the ambient air, the following rule ($r_{7}$) which is defined for each \textit{smellImplicitEvent} predicate, is also added to the program $\mathcal{P}$:

\begingroup
\fontsize{6pt}{8pt}\selectfont
  \begin{flalign*} 
&r_{7}:explained(t) :- \hspace{7pt}smellImplicitEvent(t).
  \end{flalign*}
\endgroup

Due to many reasons such as the lack of observations, or misreading of sensors, along with the aforementioned rules, we also add the two last rules ($r_{8}$ and $r_{9}$) to the cumulative part of the logic program. In this way, the answer set will always contain either the \textit{airSmellNormal} or \textit{airSmellAbnormal} as a description of the ambient air. The later, depending on the inference results, can be accompanied by the other explanations.

\begingroup
\fontsize{6pt}{8pt}\selectfont
  \begin{flalign*} 
&r_{8}:explained(t) :- \hspace{7pt}airSmellAbnormal(t).\\
&r_{9}:explained(t) :- \hspace{7pt}airSmellNormal(t).
  \end{flalign*}
\endgroup

\subsection{Generating the Volatile Logic Program ($\mathcal{Q}[t]$)}
\label{sec:programQ}

In order to guarantee having an explanation for the ambient air at each time step, the conversion process generates the volatile program $\mathcal{Q}[t]$. Given the predicate \textit{explained(t)}, the volatile part of the logic program will be modelled as given below. According to the integrity constraint rule given in the volatile part, the ASP solver needs to always provide an explanation, otherwise it ends up with dissatisfaction. For this, the solver successively accept new manifestations until the \textit{explained} predicate is inferred which consequently implies the inference of a smell.

\begingroup
\fontsize{6pt}{8pt}\selectfont
  \begin{flalign*}
&\#volatile \hspace{5pt} t.\\
&:- not \hspace{5pt} explained(t).
  \end{flalign*}
\endgroup

\begin{figure*}[t]
  \centering
    \includegraphics[width=\textwidth,height=6cm]{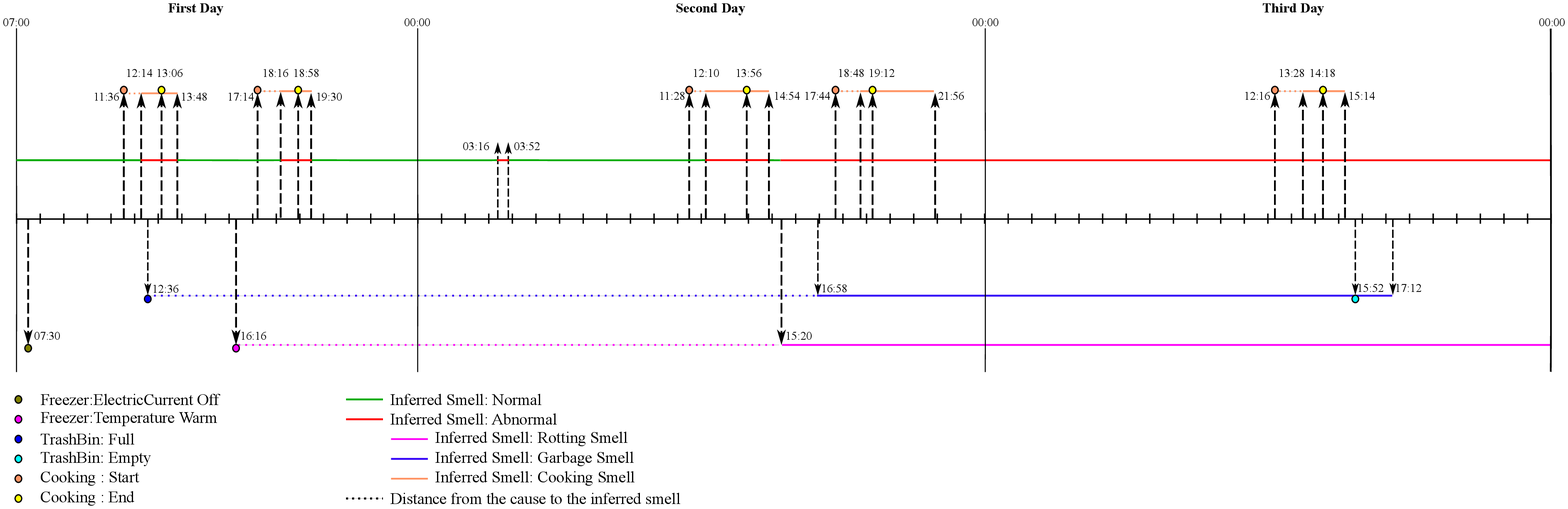}
    \caption{Set of detected events during 3-days of Observation}
    \label{fig:experiment1}
\end{figure*}

\section{Results}
\label{sec:results}

For experimental validation of the proposed method, a smart kitchen equipped with a set of sensors including a gas sensor using ZigBee wireless communication standard is deployed. Two batch measurements have been performed. The first batch is a three day run and the second batch is a five day run. The purpose of the system is to annotate each change detected by the gas sensor with an explanation that outlines the possible reasons for the change. 

Fig.~\ref{fig:experiment1} visualizes the events along with their causes for the three day experimental run. Illustrated in the legend, continuous lines are divided into two main types representing the normal (in green) and abnormal (in red) smells. As we explained in Section~\ref{sec:programP}, in case of an abnormal smell, the answer set may be augmented with inferred implicit events as the cause of the detected smell which are represented in different colors in Fig.~\ref{fig:experiment1}. The relation between a cause and its relevant inference results is also shown in dotted lines. 

\begin{table*}
 \fontsize{6pt}{7.5pt}\selectfont
 \renewcommand{\arraystretch}{1.3}
 \caption{A subset of stepwise detected manifestations along with their inferred explanations (answer sets)}
\vspace*{-1\baselineskip}
 \label{table:Experiment1}
 \begin{center}
\begin{tabular}{|c|l|l|l|l|}
  \hline
  Day & Time & TimeStep & Observation & Answer Set (Smell Description) \\
  \hline
  \hline
  \parbox[t]{2mm}{\multirow{6}{*}{\rotatebox[origin=c]{90}{First Day}}} & ... & ... & ... & ...\\
  \cline{2-5}
   & 11:36 & 16560 & manifestation(oven1, electriccurrent, on, 16560). & airSmellNormal(16560).\\
   & & & manifestation(oven1, motion, on, 16560). & \\
  \cline{2-5}
   & 12:36 & 20160 & manifestation(trashbin1, illumination, dark, 20160). & airSmellAbnormal(20160).\\
   & & & manifestation(trashbin1, door, open, 20160). & \textbf{smellCooking}(20160).\\
   \cline{2-5}
   &... &... & ... & ...\\
   \cline{2-5}
  \hline
  \parbox[t]{2mm}{\multirow{7}{*}{\rotatebox[origin=c]{90}{Second Day}}} & 03:16 & 72960 & manifestation(kitchenAir1, smell, abnormal, 72960). & airSmellAbnormal(72960).\\
  \cline{2-5}
  & ...& ...& ... & ...\\
  \cline{2-5}
  & 15:20 & 116400 & manifestation(kitchenAir1, smell, abnormal, 116400). & airSmellAbnormal(116400).\\
   & & & & \textbf{smellRotting}(116400).\\
  \cline{2-5}
  & 16:58 & 122280 & manifestation(kitchenAir1, smell, abnormal, 122280) & airSmellAbnormal(122280).\\
   & & & & \textbf{smellRotting}(122280).\\
   & & & & \textbf{smellGarbage}(122280).\\
  \cline{2-5}
  \hline
  \parbox[t]{2mm}{\multirow{6}{*}{\rotatebox[origin=c]{90}{Third Day}}} & ... & ... & ...& ...\\
   \cline{2-5}
  & 15:52 & 204720 & manifestation(trashbin1, illumination, bright, 204720).& airSmellAbnormal(204720).\\
  & & & manifestation(trashbin1, door, open, 204720).& \textbf{smellRotting}(204720).\\
   & & & & \textbf{smellGarbage}(204720).\\
  \cline{2-5}
  & 17:12 & 209520 & manifestation(kitchenAir1, smell, abnormal, 209520).& airSmellAbnormal(209520).\\
   & & & & \textbf{smellRotting}(209520).\\
   \cline{2-5}
 
  \hline
\end{tabular}		
 \end{center}
\vspace*{-0.5\baselineskip}
 \end{table*}
 
 The incremental reasoner provides the answer set for the set of manifestations given at the specific time step. The answer set expresses the smell in the environment if there is an \textit{abnormal} smell, otherwise it expresses the situation as \textit{normal} smell. For instance, at the first step ($t = 1$) the initial states of each objects' attributes in form of descriptive manifestations are given. The reasoner results in $airSmellNormal(1)$, meaning that the state of the ambient air at $t = 1$ (or first day at $7$:$00$), is \textit{normal}. On the same day, at $t = 16560$ ($11$:$36$), two manifestation indicating the oven is on and some one is moving around, are reported. According to the implicit events definition, the event \textit{Cooking} is hence inferred at $11$:$36$ (Fig.~\ref{fig:experiment1}). However, since no abnormal smell in the air is reported, the current ambient air is explained as \textit{normal}. There are many situations that due to the lack of space we cannot go through their details. Table~\ref{table:Experiment1} partially shows the answer sets of each time step considered as the explanations for the quality of the ambient air.

In order to further study the performance of the reasoner, we extended the experiments with the second package of data containing $5$ days of observations. Depending on the data package, the ontology contains different amount of individuals. Since the event based classes are time dependent, the number of their individuals depends on several parameters including the length of the observation process, the number of events happen in the environment, and the measuring unit of time in the observation process. For the three-day observation, the number of events individuals is $i \approx 762584$, and for a longer observation ($5$ days), this number increases to $i \approx 1512907$. 

\begin{figure*}[t]
  \centering
  \subfigure[Number of Individuals]
    {
        \includegraphics[width=5.5cm,height=4.7cm]{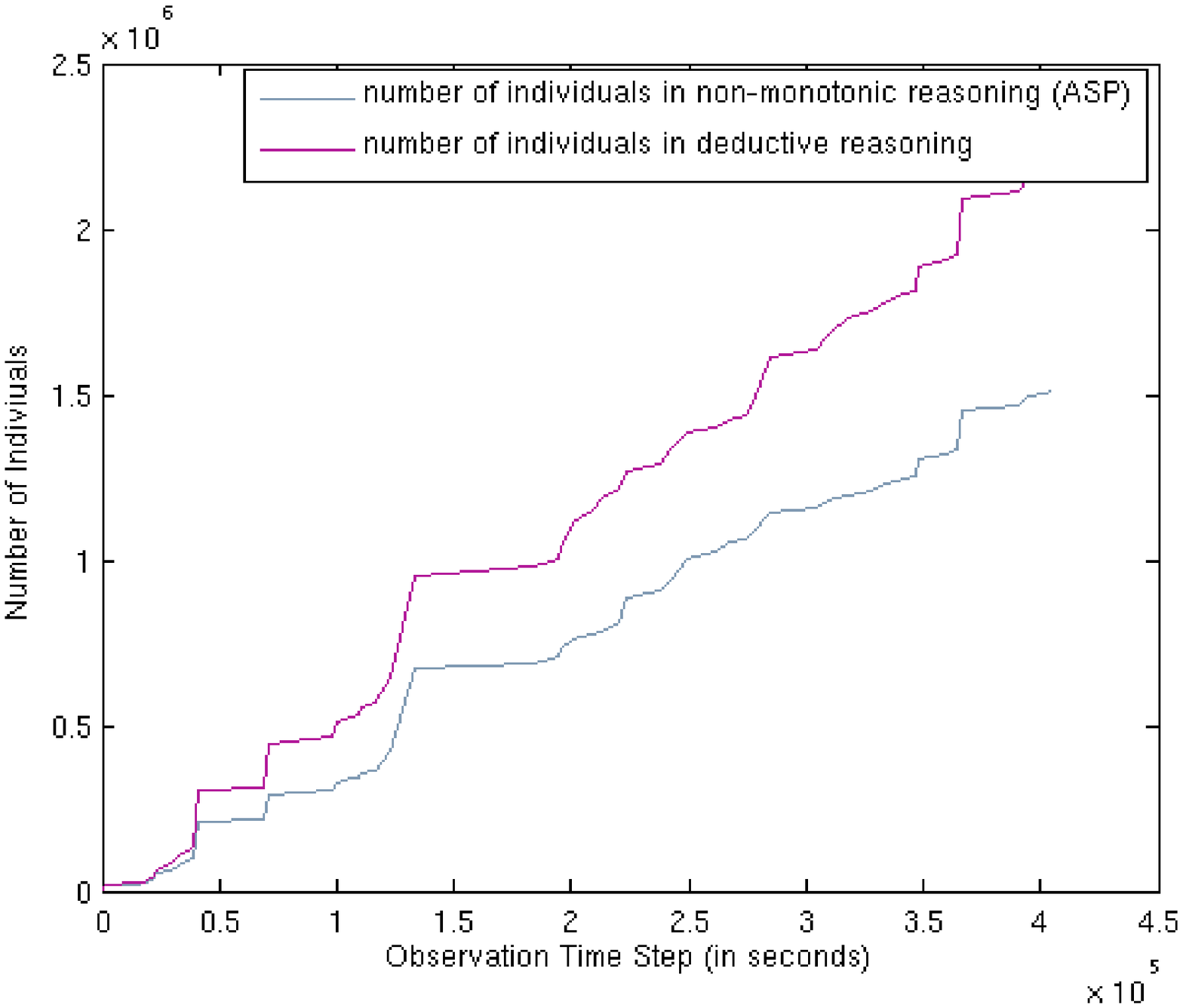}
        \label{fig:individuals}
    }
    \subfigure[Comparing Reasoning Time]
    {
        \includegraphics[width=5.5cm,height=4.5cm]{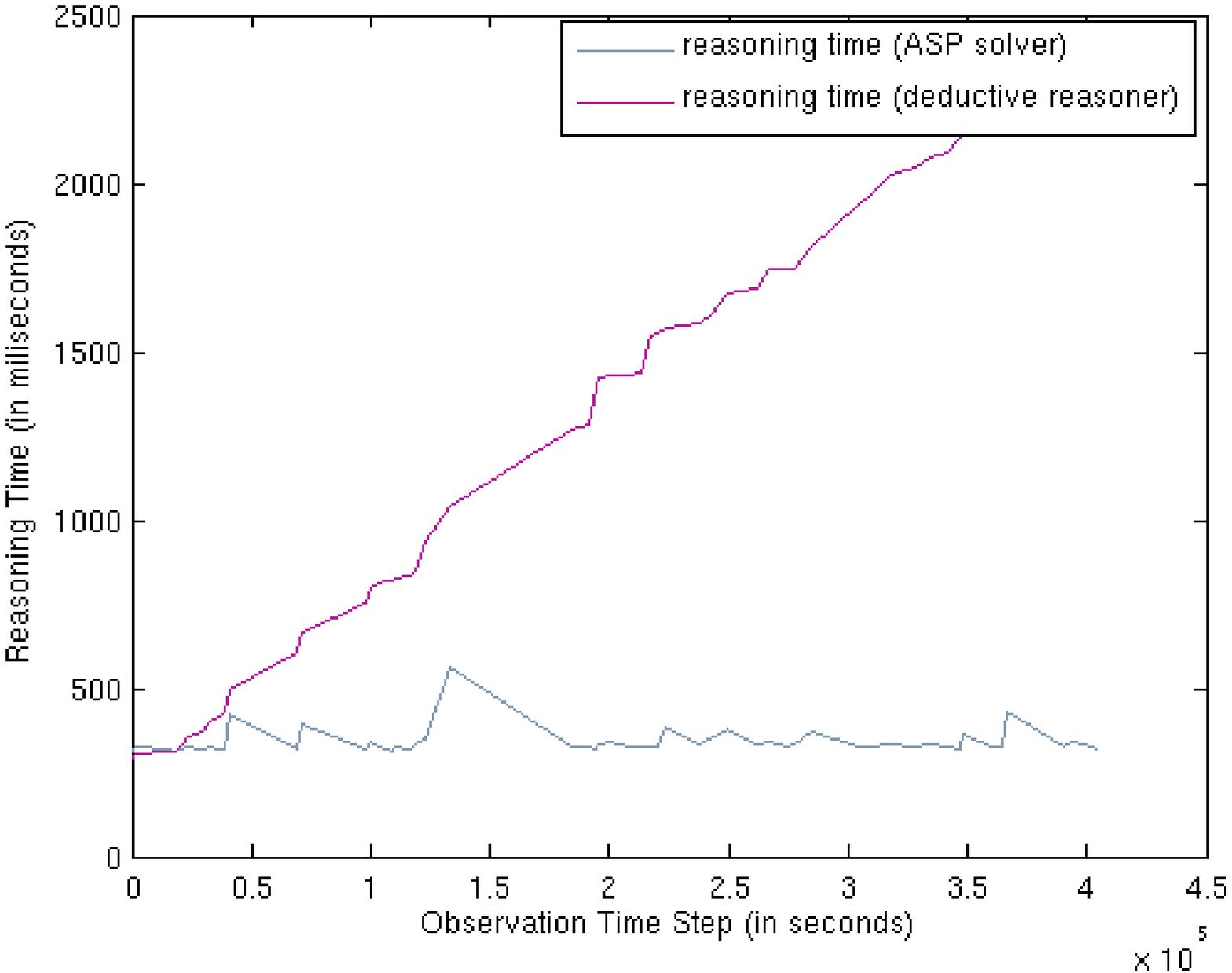}
        \label{fig:reasoningTime}
    }
  \caption{Incremental ASP vs. Pellet} 
  \label{fig:compare}
 \end{figure*}

In the following we compare the reasoning time of our system based on the incremental ASP solver with that of the monotonic ontology reasoner, Pellet \cite{A22}. Using the negation as failure (NAF) operator (which provides the closed world assumption), the reasoner results in less number of grounded individuals whereas a deductive reasoner based on open world assumption ends up with $i \approx 2174903$ individuals for the same package of data. The number of required individuals in a deductive reasoner is the ratio of the number of time steps $t$ between the two time steps at which the implicit event starts and ends ($t: t_{start}..t_{end}$). In Fig.~\ref{fig:individuals} the number of individuals increasing during the observation process is shown for both types of reasoners. The incremental solver, furthermore, extends the grounded logic program (which contains no variable but the individuals) incrementally and enables the reasoning process to only consider the recently added manifestations (events) during the solving process, rather than the entire grounded individuals. Figure.~\ref{fig:reasoningTime} shows the difference of the reasoning time between the deductive Pellet ontology reasoner and the ASP incremental solver. Therefore, in addition of the re-usability feature, our ontology-based knowledge representation and reasoning approach provides a considerable efficiency in reasoning time which consequently enhances the scalability of the system.
 
\section{Conclusion}
\label{sec:conclusion}

The key point in this work is enabling non-monotonic reasoning over an ontology which is inspired from the SSN ontology. 
Integrating the ontology with the ASP solver provides the opportunity of intuitively modelling the environment at different phases of the process. The incremental ASP solver also enables the encoding of the environment's history incrementally which leads into an efficient reasoning time. The ASP semantics, in addition, simplifies the automated creation process of the logic program. The whole logic program except the implicit events creation is done automatically. The manually added rules related to implicit events are also defined based on the combination of the existing expressive predicates that state explicit events. 

However, the temporal relation between manifestations can be further developed. If the number of states of an object increases, the number of rules considering temporal relation between  events and their causes also grows. Extending the temporal relations between events, as our future work, can leads to a more enriched temporal reasoning.


\begin{thebibliography}{4}
   
 \bibitem{M2}
M.~Compton and P.~Barnaghi and L.~Bermudez and R.~Garcia-Castro and O.~Corcho and S.~Cox and J.~Graybeal and M.~Hauswirth and C.~Henson and A.~Herzog and V.~Huang and K.~Janowicz and W.~Kelsey and D.~Phuoc and L.~Lefort and M.~Leggieri and H.~Neuhaus and A.~Nikolov and K.~Page and A.~Passant and A.~Sheth and K.~Taylor, 
  \emph{The SSN ontology of the W3C semantic sensor network incubator group}.
  J.Web Semantics: Science, Services and Agents on the World Wide Web
  (17), P.25-32,
  Elsevier,
  2012.
  
   
  \bibitem{P10}
  W.~Wang and P.~Barnaghi,
  \emph{Semantic Annotation and Reasoning for Sensor Data}.
  Lecture Notes in Computer Science,
  P.66-76,
  Springer,
  2009.  
  

   
  \bibitem{P17}
  M.~Gebser and T.~Grote and R.~Kaminski and Ph.~Obermeier and O.~Sabuncu and T.~Schaub,
  \emph{Stream Reasoning with Answer Set Programming: Preliminary Report}.
  Proc. 13th Int. Conf. on the Principles of Knowledge Representation and Reasoning (KR),
  AAAI Press,
  2012. 
  
  \bibitem{P40}
  A.~Artikis and M.~Sergot and G.~Paliouras,
  \emph{A Logic Programming Approach to Activity Recognition}.
  Proc. 2Nd ACM International Workshop on Events in Multimedia,
  P.3-8,
  ACM,
  2010.
  
  \bibitem{A15}
  A.~Mileo and D.~Merico and S.~Pinardi and R.~Bisiani,
  \emph{A Logical Approach to Home Healthcare with Intelligent Sensor-Network Support}.
  The Computer Journal,
  P.1257-1276,
  Oxford University Press,
  2010.
  
   
  

\bibitem{A22} 
E.~Sirin and B.~Parsia and B.C.~Grau and A.~Kalyanpur and Y.~Katz,
\emph{Pellet: A Practical OWL-DL Reasoner},
Web Semant. 
(5), P.51-53,
Elsevier Science Publishers B. V.,
2007.


\end{thebibliography}
\end{document}